\documentclass[cameraready]{Interspeech}
\usepackage{multirow}
\usepackage{comment}
\usepackage{amssymb}

\newcommand{\cursym}{\mbox{\scalebox{0.75}{${\dagger}$}}}
\newcommand{\curaff}{\mbox{\scalebox{0.75}{${\ddagger}$}}}
\newcommand{\curaffb}{\mbox{\scalebox{0.75}{${\lozenge}$}}}

\title{To Be Multimodal or Not to Be:\\Query-Adaptive Audio-Visual Person Retrieval via Active Modality Detection}

\author[affiliation={1,\cursym}]{Erfan}{Loweimi}
\author[affiliation={1}]{Mengjie}{Qian}
\author[affiliation={1}]{Kate}{Knill}
\author[affiliation={2,\curaff}]{Guanfeng}{Wu}
\author[affiliation={3}]{Chi-Ho}{Chan}
\author[affiliation={2,\curaffb}]{Abbas}{Haider}
\author[affiliation={3}]{Muhammad}{Awan}
\author[affiliation={3}]{Josef}{Kittler}
\author[affiliation={2}]{Hui}{Wang}
\author[affiliation={1}]{Mark}{Gales}

\address{
    $^1$University of Cambridge, UK; $^2$Queen's University Belfast, UK; $^3$University of Surrey, UK\\
    $^{\dagger}$Cisco, UK; $^\ddagger$Southwest Jiaotong University, China; $^{\lozenge}$Teesside University, UK
}

\email{}


\begin{document}

\maketitle

\begin{abstract}
When retrieving a person from a video archive by voice and face, should the system \emph{be multimodal or not}? In real-world broadcast archives, unlike curated benchmarks, a target may be heard but unseen, seen but unheard, or both. Fusing scores from an absent modality injects noise, degrading precision below the best unimodal system. We propose a query-adaptive framework that detects active modalities via cross-modal score consistency: when both modalities are active, files retrieved by one also score highly on the other; this agreement breaks down when a modality is absent. Classifiers driven by these cross-modal features achieve 89\% detection accuracy. On the BBC Rewind corpus (with over 12,000 broadcast videos) the adaptive system attains 94.2\% P@1, outperforming speaker-only (82.9\%), face-only (93.4\%), and fixed fusion (90.0\%), recovering 64\% of the gap to an oracle with ground-truth modality labels (96.6\%).
\end{abstract}

% ===========================================================================
\section{Introduction}
\label{sec:intro}

Locating a specific individual across a large-scale video archive is critical for journalism, forensics, and media indexing~\cite{Stone2014,Ruger2010,Larson2012,Chen2023}. Person retrieval can exploit two complementary biometric modalities: speaker voice via speaker embeddings~\cite{Snyder2018,Desplanques2020,Koluguri2021} and facial appearance via face embeddings~\cite{Schroff2015,Deng2019arcface}. When both modalities are available, a natural question arises: should the retrieval \emph{be multimodal}, i.e., combining voice and face evidence or not?
Standard benchmarks such as VoxCeleb~\cite{Nagrani2017,Chung2018voxceleb2} are purpose-built: subjects are carefully selected and each clip is curated to ensure the target is both seen and heard. Real-world broadcast archives, by contrast, are not collected with any retrieval task in mind. The BBC Rewind corpus~\cite{BBCRewind2024}, for instance, comprises decades of journalistic footage (e.g., interviews, parliamentary debates, street reports, voice-over narrations) where a person of interest may contribute only their voice, only their face, or both, depending entirely on the editorial context.

Analysis of this corpus in prior work~\cite{MVSE,Loweimi2024} revealed three distinct \emph{presence types}:
\begin{itemize}
    \item \textit{Audio-Visual Presence (AVP):} the person is both seen and heard---the ideal multimodal case;
    \item \textit{Audio-only Presence (AoP):} the person is heard but not visible, e.g., a voice-over narrator or off-camera interviewer;
    \item \textit{Visual-only Presence (VoP):} the person is visible but does not speak, e.g., appearing in footage without talking to the mic.
\end{itemize}
When only one modality is active, the embedding of the absent modality encodes an \emph{unrelated} identity. For example, in a VoP query, the speaker embedding captures an arbitrary voice rather than the target, and fusing this misleading score with the informative face score distorts the ranking. This makes fixed-weight fusion suboptimal and \emph{worse} than the best unimodal system.

Prior work on audio-visual fusion~\cite{Nagrani2018seeing,Afouras2020,Baltrusaitis2019} has largely assumed that both modalities are simultaneously informative. Baltru\v{s}aitis et al.~\cite{Baltrusaitis2019} surveyed early, late, and hybrid fusion strategies, noting that late fusion offers robustness when modality quality varies, but did not consider the case where a modality is entirely uninformative or absent. The \textit{multimodal video search by examples} (MVSE) framework~\cite{MVSE,Wu2024,Qian2024} demonstrated that multimodal fusion outperforms single-modal retrieval on BBC Rewind and that audio quality impacts precision more severely than visual quality~\cite{Wu2024}, but assumed both modalities are always available. Unimodal speaker retrieval was studied in~\cite{Loweimi2024,Loweimi2025spcom}, benchmarking x-vector~\cite{Snyder2018}, ECAPA-TDNN~\cite{Desplanques2020}, and TitaNet~\cite{Koluguri2021} embeddings and revealing challenges of noisy metadata labels and acoustic diversity of the recordings. None of these works address the scenario where one modality may be entirely absent or uninformative.

This paper addresses the \emph{modality-absence problem} through a query-adaptive framework that (1)~detects the active modalities for each query by analysing inter-modal consistency of retrieval score distributions, and (2)~adapts the fusion weight accordingly. The key insight is that cross-modal scores---evaluating one modality's retrieval set through the lens of the other---provide a natural diagnostic for modality presence: high cross-modal agreement signals that both modalities are active, while low agreement exposes an absent modality. Building on this, our contributions are: a novel feature design combining within-modal and cross-modal cosine similarity scores for presence-type classification; rigorous evaluation under multiple cross-validation protocols; and demonstration that adaptive fusion significantly outperforms all baselines (89\% detection accuracy, 94.2\% P@1, recovering the gap to an oracle).

The rest of this paper is organised as follows. Section~\ref{sec:mvse} reviews the MVSE framework, including its speaker and face embedding pipelines. Section~\ref{sec:framework} presents the proposed query-adaptive scoring, fusion, and active modality detection. Section~\ref{sec:setup} describes the experimental setup, and Section~\ref{sec:results} presents and discusses the results. Finally, Section~\ref{sec:conclusion} concludes the paper.

% ===========================================================================
\section{The MVSE Framework}
\label{sec:mvse}

This work builds upon the Multimodal Video Search by Examples (MVSE) framework~\cite{MVSE,Wu2024}, an EPSRC-funded system for content-based retrieval in the BBC Rewind archive---a publicly available collection of 12,594 video files (409\,h) spanning 1948--1979, covering news footage with diverse acoustic and visual conditions~\cite{BBCRewind2024}. %Person names are extracted from journalist-curated synopses via named entity recognition~\cite{spacy2015}, yielding ${\sim}$5,800 distinct identities. Crucially, these synopses were written for editorial purposes, not for retrieval---names may be misspelt, incomplete, or refer to individuals not actually appearing in the video~\cite{Loweimi2025spcom}.
Figure~\ref{fig:workflow} illustrates the MVSE pipeline.%, whose two person-identity modules are summarised below.

The speaker embedding extraction module begins with speaker diarisation via PyAnnote~\cite{Bredin2020,Bredin2021} to segment each video into per-speaker regions. Speaker embeddings are then extracted with a pre-trained ECAPA-TDNN~\cite{Desplanques2020} from SpeechBrain~\cite{Ravanelli2021ecapa}, selected over x-vectors~\cite{Snyder2018} and TitaNet~\cite{Koluguri2021} based on benchmarking in~\cite{Loweimi2024,Loweimi2025spcom}. The model is trained on VoxCeleb~1\&2~\cite{Nagrani2017,Nagrani2020} ($>$2,000\,h, 7,205 speakers) with AM-Softmax loss~\cite{Wang2018amsoftmax}, achieving 0.8\% EER on the VoxCeleb test set~\cite{Wu2024}. Segment-level embeddings are aggregated into per-speaker representations via \emph{duration-weighted averaging}~\cite{Wu2024,Loweimi2025spcom}, which assigns higher importance to longer, more informative segments, yielding one embedding per speaker per video.

The face embedding extraction module detects faces per frame using the heatmap-assisted detector of~\cite{Ju2023} and geometrically normalises them via Umeyama transforms~\cite{Umeyama1991}. Face embeddings are extracted with a ResNet-400 backbone trained on WebFace42M~\cite{Zhu2021webface}. %(99.85\% accuracy on LFW)~\cite{Wu2024}.
Non-frontal faces and those smaller than 45\,px are discarded. Cosine-angle clustering groups per-frame embeddings into per-identity clusters, yielding one representative face embedding per person per video~\cite{Wu2024}.

All embedding models are used \emph{zero-shot}: the noisy metadata-derived labels in the corpus make supervised adaptation unreliable~\cite{Loweimi2024,Loweimi2025spcom}. Given a query video, each module compares the query embedding against the pre-indexed archive embeddings via cosine similarity, producing a per-modality score for every archive file. These scores are then combined through score-level fusion, and the top-$n$ files are returned.% as a ranked list.

Wu et al.~\cite{Wu2024} extensively analysed BBC Rewind data in the context of MVSE project and demonstrated multimodal fusion consistently outperforms single-modal retrieval. %(2)~audio quality impacts precision more severely than visual quality, indicating the speaker modality is more sensitive to recording conditions; and
%(2)~cosine similarity outperforms Hamming distance.
However, the framework treats all modalities equally during fusion, assuming each is informative for every query. When a person is only heard or only seen, the uninformative modality injects noise, degrading retrieval. The present work addresses this limitation by introducing a modality detection module that determines, per query, which modalities are active \emph{before} fusion is applied.
% ===========================================================================
\section{Query-Adaptive Retrieval Framework}
\label{sec:framework}

\begin{figure}[t]
  \centering
  \includegraphics[width=\linewidth]{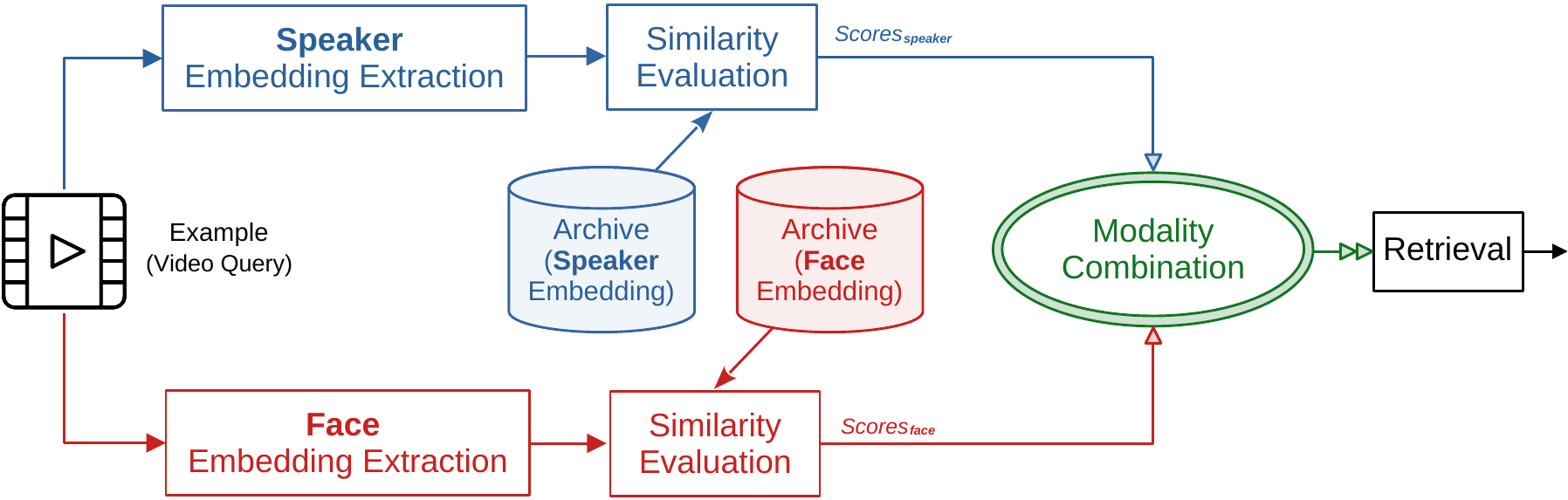}
  \caption{Query-adaptive MVSE framework for multimodal person retrieval. The modality combination module analyses modality scores to decide whether to be multimodal or not.}
  \label{fig:workflow}
\end{figure}

Figure~\ref{fig:workflow} illustrates the proposed extension to the MVSE pipeline: a \textit{modality combination} module that detects which modalities are active for a given query and sets the fusion weight accordingly, before producing the final ranked list.

\subsection{Scoring and fusion}
\label{ssec:scoring}

Given query embeddings $\mathbf{e}_{\text{spk}}^{(q)}$ and $\mathbf{e}_{\text{face}}^{(q)}$, the per-modality scores for each archive video file $\text{ID}_i$ are:
\begin{align}
    s_{\text{spk}}[\text{ID}_i] &= \max_{1 \le j \le M_i} \cos(\mathbf{e}_{\text{spk}}^{(q)},\, \mathbf{e}_{\text{spk}}^{(i,j)}) \label{eq:score_spk} \\
    s_{\text{face}}[\text{ID}_i] &= \max_{1 \le k \le F_i} \cos(\mathbf{e}_{\text{face}}^{(q)},\, \mathbf{e}_{\text{face}}^{(i,k)}) \label{eq:score_face}
\end{align}
\noindent where $\cos(\cdot,\cdot)$ denotes cosine similarity, $\mathbf{e}_{\text{spk}}^{(i,j)}$ and $\mathbf{e}_{\text{face}}^{(i,k)}$ are the $j$-th speaker and $k$-th face embeddings in the archive file~$i$, and $M_i$ and $F_i$ are the numbers of detected speakers and faces in that file. The $\max$ selects the best-matching identity per file. The fused multimodal score ($s_{\text{MM}}$) is:
\begin{align}
    s_{\text{MM}}[\text{ID}_i] = \lambda\,s_{\text{spk}}[\text{ID}_i] + (1{-}\lambda)\,s_{\text{face}}[\text{ID}_i]
    \label{eq:mm_score}
\end{align}
\noindent where $\lambda \in [0,1]$ is the fusion weight controlling the relative contribution of each modality. We adopt late (score-level) fusion for its modularity and because it permits direct control of~$\lambda$, which is the key parameter in our adaptive framework. Files are ranked by $s_{\text{MM}}$ in descending order.

The optimal $\lambda$ is \emph{query-dependent}: $\lambda{=}1$ for AoP, $\lambda{=}0$ for VoP, and $\lambda{\approx}0.5$ for AVP. To understand why a fixed $\lambda$ fails, consider a VoP query: the speaker embedding captures an arbitrary voice (not the target), so $s_{\text{spk}}$ is drawn from the distribution of inter-speaker similarities across the archive. This noise term has non-negligible variance, meaning different archive files receive different spurious speaker scores, creating rank perturbations that push the correct file below incorrect ones. The problem is symmetric for AoP, where irrelevant face scores contaminate fusion. As shown in Section~\ref{sec:results}, this effect makes fixed fusion \emph{worse} than the best unimodal system at P@1.

\subsection{Active modality detection}
\label{ssec:modality_detection}

The modality detection module classifies each query into \{AoP, VoP, AVP\} and sets~$\lambda$ accordingly ($\lambda{=}1$, $0$, or $0.5$).

\subsubsection{Feature design}

\begin{figure}[t]
  \centering
  \includegraphics[width=\linewidth,height=51mm]{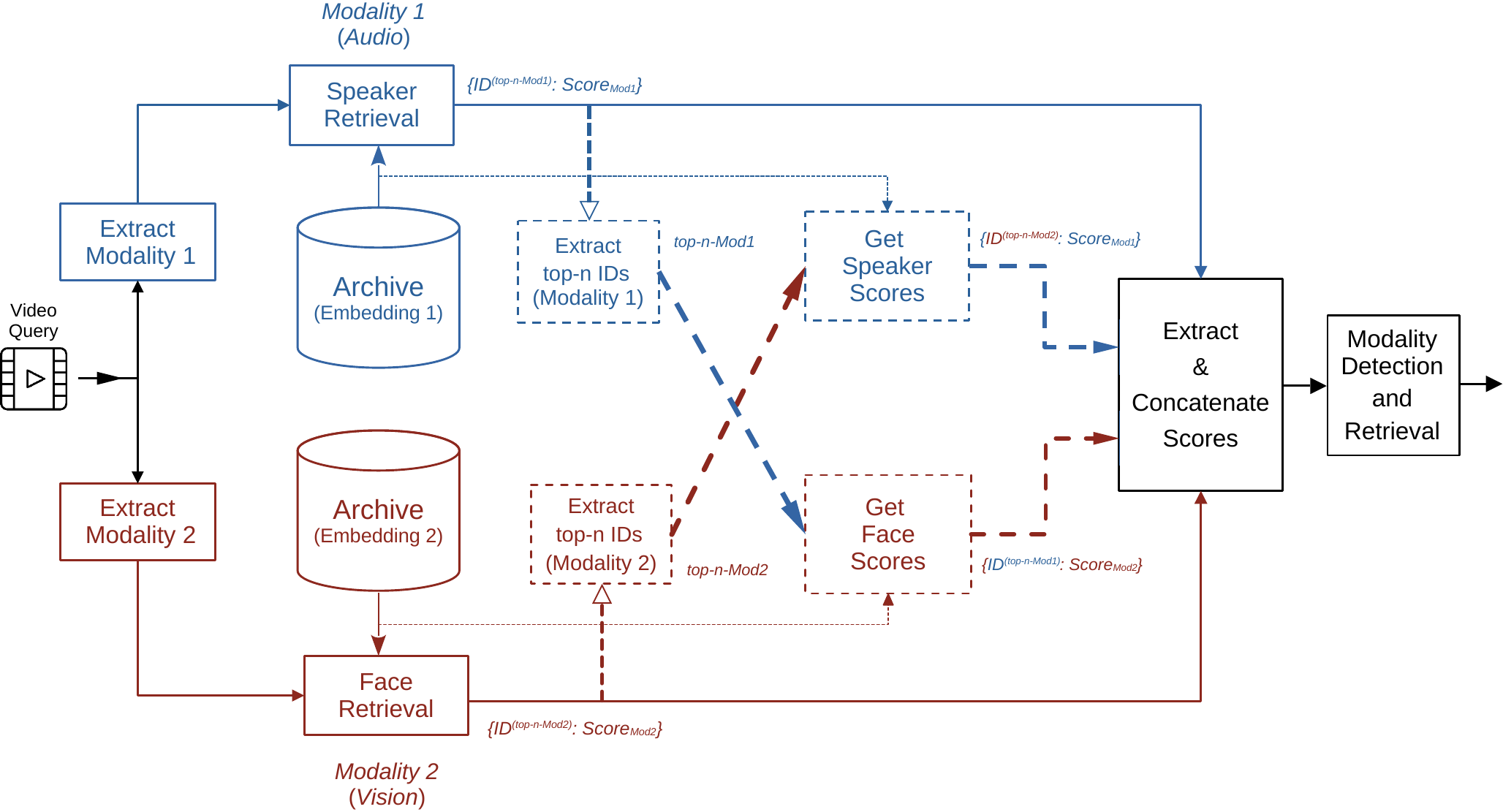}
  \caption{Feature extraction for modality detection. Solid lines: within-modal scores; dashed lines: cross-modal scores.} %High cross-modal consistency indicates AVP; low consistency signals a single active modality.}
  \label{fig:modality-class}
\end{figure}

For each query, we retrieve the top-$n$ files per modality and extract features from the resulting score distributions (Figure~\ref{fig:modality-class}).

\textbf{Within-modal scores.} Let $\mathcal{R}_s^{n}$ and $\mathcal{R}_f^{n}$ denote the top-$n$ archive files retrieved by the speaker and face modalities for query $i$. The within-modal score vectors are:
\begin{align}
    \mathbf{s}_s &= \bigl[s_{\text{spk}}(i)\bigr]_{i \in \mathcal{R}_s^{n}}, \quad
    \mathbf{s}_f = \bigl[s_{\text{face}}(i)\bigr]_{i \in \mathcal{R}_f^{n}}
    \label{eq:within}
\end{align}
\noindent where $\mathbf{s}_s \in \mathbb{R}^{n}$ and $\mathbf{s}_f \in \mathbb{R}^{n}$ are the speaker and face score vectors, comprising the cosine similarity scores of the top-$n$ retrieved files for the respective modality, sorted in descending order. 
An active modality typically produces a \emph{peaked} distribution, in which the top files containing the target person receive high similarity scores followed by a sharp drop-off. In contrast, an inactive modality yields a \emph{flatter} distribution with lower overall magnitude, as the query embedding encodes an unrelated identity with no genuine matches. However, the within-modal distribution alone is insufficient for reliable detection. Some speakers exhibit naturally lower inter-speaker discriminability due to acoustic similarity or poor recording conditions, resulting in ambiguous distributions even when the modality is active.

\textbf{Cross-modal scores.} We evaluate each modality's scores on the retrieval set of the \emph{other} modality:
\begin{align}
    \mathbf{c}_{s \!\to\! f} &= \bigl[s_{\text{face}}(i)\bigr]_{i \in \mathcal{R}_s^{n}}, \quad
    \mathbf{c}_{f \!\to\! s} = \bigl[s_{\text{spk}}(i)\bigr]_{i \in \mathcal{R}_f^{n}}
    \label{eq:cross}
\end{align}
\noindent where $\mathbf{c}_{s \!\to\! f} \in \mathbb{R}^{n}$ contains the \emph{face} scores of the files retrieved by the \emph{speaker} modality, and $\mathbf{c}_{f \!\to\! s} \in \mathbb{R}^{n}$ contains the \emph{speaker} scores of the files retrieved by the \emph{face} modality. These capture \emph{inter-modal consistency}: for AVP queries, the files retrieved by speaker similarity should also contain the target face (and vice versa), yielding high cross-modal scores. For AoP queries, the speaker-retrieved files are relevant, but these files need not contain the target face (since the person is not visually present), so $\mathbf{c}_{s \!\to\! f}$ will be low. The pattern is symmetric for VoP.

This inter-modal consistency is the central discriminative signal: as we will see in the experimental results (Table~\ref{tab:class_acc}), adding cross-modal features yields an ${\sim}$6\,percentage point (pp) gain in classification accuracy, confirming that within-modal score distributions alone are insufficient.

\textbf{Summary statistics.} The mean and standard deviation of each score vector are appended by
\begin{align}
\boldsymbol{\mu}&=[\mu(\mathbf{s}_s),\,\mu(\mathbf{s}_f),\,\mu(\mathbf{c}_{s \!\to\! f}),\,\mu(\mathbf{c}_{f \!\to\! s})]\\
\boldsymbol{\sigma}&=[\sigma(\mathbf{s}_s),\,\sigma(\mathbf{s}_f),\,\sigma(\mathbf{c}_{s \!\to\! f}),\,\sigma(\mathbf{c}_{f \!\to\! s})]
\end{align}
yielding the full feature vector $\mathbf{f} \in \mathbb{R}^{4n+8}$:
\begin{align}
    \mathbf{f} = [\,\mathbf{s}_s;\, \mathbf{s}_f;\, \mathbf{c}_{s \!\to\! f};\, \mathbf{c}_{f \!\to\! s};\, \boldsymbol{\mu};\, \boldsymbol{\sigma}\,]
    \label{eq:feat_full}
\end{align}
\noindent where $[\,\cdot\,;\,\cdot\,]$ denotes concatenation, $\boldsymbol{\mu} \in \mathbb{R}^{4}$ and $\boldsymbol{\sigma} \in \mathbb{R}^{4}$ are the means and standard deviations of the four score vectors.%, and $\mathbf{f} \in \mathbb{R}^{4n+8}$.

Table~\ref{tab:expected} summarises the expected behaviour of $\boldsymbol{\mu}$ and $\boldsymbol{\sigma}$: an active modality yields high mean and low variance (genuine matches cluster tightly), whereas an inactive modality yields low mean and high variance (scores are drawn from the tail of the impostor distribution). These patterns provide a principled basis for the feature design. 
Rather than hand-crafting rules from these expected patterns (e.g., thresholding mean or variance), we adopt a data-driven approach: classifiers learn the decision boundaries from labelled examples, capturing interactions across all score vectors and generalising more robustly to unseen speakers and conditions.

The choice of $n$ reflects a trade-off: larger $n$ yields richer feature vectors, but the core assumption, that all top-$n$ scores correspond to the target person, requires the person to appear in at least $n$ archive files. If a person appears in fewer than $n$ files, the remaining scores belong to impostors, contaminating the feature vector. We set $n{=}10$ as the largest value for which this assumption holds across all query speakers.

\begin{table}[t]
  \caption{Expected score statistics by presence type. An active modality produces high mean and low standard deviation (SD); an inactive modality exhibits the opposite pattern.}
  \label{tab:expected}
  \centering
  \begin{tabular}{l c c c c}
    \toprule
    & \multicolumn{2}{c}{\textbf{Mean}} & \multicolumn{2}{c}{\textbf{SD}} \\
    \cmidrule(lr){2-3} \cmidrule(lr){4-5}
    \textbf{Type} & \textbf{Spk} & \textbf{Face} & \textbf{Spk} & \textbf{Face} \\
    \midrule
    AVP & High & High & Low  & Low  \\
    AoP & High & Low  & Low  & High \\
    VoP & Low  & High & High & Low  \\
    \bottomrule
  \end{tabular}
\end{table}

\subsubsection{Classifiers}

Given the limited labelled query data (523 video files, 21.1 hours), which reflects the high effort required to label audio-visual presence types reliably in real-world broadcast material, and the low feature dimensionality ($4n{+}8{=}48$ for $n{=}10$), we employ logistic regression (LogReg), SVMs with linear (SVM-L) and RBF (SVM-R) kernels, and decision trees (DT)~\cite{scikit-learn}. %The consistency of results across classifiers (Table~\ref{tab:class_acc}) confirms a relatively simple classification boundary.

% ===========================================================================
\section{Experimental Setup}
\label{sec:setup}

The BBC Rewind corpus~\cite{BBCRewind2024} is a publicly available, in-the-wild broadcast archive from Northern Ireland, comprising 12,594 video files (409\,hours) spanning 1948--1979. Unlike standard curated academic datasets, BBC Rewind reflects real editorial footage, including interviews, debates, voice-overs, and crowd scenes, where a person may be seen, heard, both, or neither. Person names are obtained by applying named entity recognition (NER)~\cite{spacy2015} to journalist-written synopses, yielding ${\sim}$5,800 distinct names. The archive poses substantial audio-visual challenges, including variable recording technology across decades, background noise and speaker overlap, black-and-white and low-resolution video, motion blur, and pose/occlusion, making it a valuable testbed for real-world retrieval and media indexing.

Our query set comprises 523 video files (21.1\,hours) from 38 prominent politicians. Manual verification~\cite{Loweimi2025spcom} identified 425 audio-visual presence (AVP) queries, 72 visual-only presence (VoP; face visible but no speech), and 26 audio-only presence (AoP; speech only).

For retrieval evaluation, we report Precision@$K$ (P@$K$) for $K{\in}\{1,3,5,10\}$~\cite{Manning2009}. A retrieved file is considered relevant if the query person’s name appears in its synopsis, which was validated as a reliable proxy for these prominent figures through extensive analysis in~\cite{Loweimi2025spcom}. For modality detection, we report accuracy under leave-one-speaker-out cross validation (LoSoCV). We focus on LoSoCV instead of vanilla k-fold cross validation, as it is the most practically relevant setting and tests whether the classifier generalises to entirely unseen speakers.

% ===========================================================================
\begin{table}[t]
  \caption{Modality classification accuracy (\%) under LoSoCV. ``Cross'' = cross-modal scores; ``$\mu+\sigma$'' = summary statistics.}
  \label{tab:class_acc}
  \centering
  \begin{tabular}{l c c c c}
    \toprule
    \textbf{Features} & \textbf{LogReg} & \textbf{SVM-L} & \textbf{SVM-R} & \textbf{DT} \\
    \midrule
    Base              & 82.3 & 82.8 & 82.7 & 76.7 \\
    +\,Cross          & 88.2 & 88.1 & 87.9 & 88.8 \\
    +\,Cross+$\mu$+$\sigma$ & \textbf{88.5} & \textbf{88.4} & \textbf{88.2} & \textbf{89.1} \\
    \bottomrule
  \end{tabular}
\end{table}

\section{Results and Discussion}
\label{sec:results}

\subsection{Modality classification}
Table~\ref{tab:class_acc} reports classification accuracy under LoSoCV.
Within-modal scores alone achieve ${\sim}$82\%, already well above the 81.3\% majority-class baseline (note AVP accounts for 425/523 queries). Adding cross-modal scores yields a ${\sim}$6\,pp boost, confirming inter-modal consistency as the dominant discriminative signal for modality detection. Summary statistics ($\boldsymbol{\mu}$, $\boldsymbol{\sigma}$) provide a marginal further gain ($<$0.5\,pp), indicating that the raw score vectors already capture the distributional shape information that mean and standard deviation summarise.

Performance is remarkably consistent across classifiers (within ${\sim}$1\,pp), with the decision tree achieving the best accuracy of 89.1\%. This consistency suggests a relatively clean, near-linearly separable classification boundary in the feature space---the three presence types occupy distinct regions characterised by different patterns of cross-modal correlation. The high variance across speakers under LoSoCV (standard deviations of 13--20\%) reflects that some speakers appear in more challenging conditions (lower SNR, farther from the mic, fewer archive files, atypical editorial framing) than others, making their presence type harder to classify. Notably, LoSoCV is the most stringent protocol: it tests generalisation to speakers entirely unseen during training, simulating realistic deployment where new query identities continually arrive.

\subsection{Retrieval performance}

Table~\ref{tab:retrieval} shows retrieval results. We discuss them case by case.% on $\mathcal{Q}^\dagger$.

\begin{table}[t]
  \caption{Person retrieval performance (P@$K$, \%). ``Fixed'' uses $\lambda{=}0.5$. ``Adaptive'' uses modality detection.} %``Oracle'' uses ground-truth labels.}
  \label{tab:retrieval}
  \centering
  \begin{tabular}{l l c c c c}
    \toprule
    \textbf{System} & \textbf{Features} & \textbf{P@1} & \textbf{P@3} & \textbf{P@5} & \textbf{P@10} \\
    \midrule
    Speaker  & -- & 82.9 & 80.7 & 78.3 & 74.3 \\
    Face     & -- & 93.4 & 88.6 & 86.3 & 81.6 \\
    \midrule
    Fixed  & -- & 90.0 & 88.6 & 87.0 & 83.3 \\
    \midrule
    Adaptive     & Base       & 92.1 & 88.8 & 86.8 & 82.8 \\
    Adaptive     & +Cross     & 94.2 & 90.4 & 88.0 & 84.1 \\
    Adaptive     & +Cross+$\mu$+$\sigma$ & \textbf{94.2} & \textbf{90.6} & \textbf{88.2} & \textbf{84.2} \\
    \midrule
    Oracle & -- & 96.6 & 91.8 & 89.3 & 85.2 \\
    \bottomrule
  \end{tabular}
\end{table}

\textbf{Unimodal baselines.} Face retrieval (93.4\% P@1) outperforms speaker retrieval (82.9\%) by 10.5\,pp. This gap is largely driven by the audio modality's susceptibility to broadcast degradations, including background noise, reverberation, and overlapping speech. In contrast, cameras can zoom in on the target person to isolate the subject, yielding relatively clean facial captures even in complex scenes. This asymmetry implies that fixed fusion weights are suboptimal, as the relative reliability of each modality varies across the archive.

\textbf{Fixed fusion degrades P@1.} We set $\lambda{=}0.5$ to weight both modalities equally and avoid systematic bias. Despite this, fixed fusion yields a P@1 of 90.0\%, falling below the face-only baseline (93.4\%). This finding highlights the central empirical motivation of this work: naively enforcing multimodality when one modality is absent proves counterproductive, as spurious scores from the inactive modality corrupt the ranking.

\textbf{Adaptive fusion.} The adaptive system achieves 94.2\% P@1, outperforming all baselines with consistent gains across all values of $K$. Cross-modal features account for a 2.1\,pp improvement over Base features (92.1\% $\to$ 94.2\%), mirroring the 6\,pp classification gain in Table~\ref{tab:class_acc}. This strong correspondence confirms that retrieval performance is directly driven by modality detection accuracy, as each percentage point of classification improvement translates to measurable retrieval gains.

\textbf{Oracle.} With ground-truth presence-type labels, P@1 reaches 96.6\%. The adaptive system recovers $\frac{94.2-90.0}{96.6-90.0}{=}64\%$ of the gap from fixed fusion to the oracle. The remaining gap to 100\% (3.4\,pp) reflects inherent limitations of the underlying embedding models and the noisy synopsis-based evaluation.% criterion.

\begin{table}[t]
  \caption{P@1 (\%) by presence type. Fixed: $\lambda{=}0.5$.}
  \label{tab:per_type}
  \centering
  \begin{tabular}{l c c c}
    \toprule
    \textbf{System} & \textbf{AVP} & \textbf{AoP} & \textbf{VoP} \\
    \midrule
    Speaker     & 86.6 & 80.8 & -- \\
    Face        & 95.1 & -- & 93.4 \\
    \midrule
    Fixed       & 93.8 & 76.9 & 88.5 \\
    \midrule
    Adaptive    & \textbf{95.5} & \textbf{80.8} & \textbf{93.4} \\
    \midrule
    Oracle      & 96.9 & 80.8 & 93.4 \\
    \bottomrule
  \end{tabular}
  \vspace{-1mm}
\end{table}

\subsection{Per-presence-type analysis}

Table~\ref{tab:per_type} decomposes P@1 by presence type.
Fixed fusion degrades AoP by 3.9\,pp (from 80.8\% to 76.9\%) and VoP by 4.9\,pp (from 93.4\% to 88.5\%) relative to the respective unimodal baselines. %---a substantial penalty given that these queries constitute 18.8\% of the evaluation set ($\mathcal{Q}^\dagger$).
The adaptive system recovers full unimodal performance for both types, matching the oracle. This indicates near-perfect detection of AoP and VoP queries, which can be attributed to the strong discriminative power of cross-modal scores: when a modality is entirely absent, there is no correlation between the retrieval sets of the two modalities, making the cross-modal score vector a reliable indicator.

For AVP, the adaptive system achieves 95.5\%, outperforming both unimodal systems (86.6\% speaker, 95.1\% face). The 0.4\,pp gain over face-only demonstrates genuine multimodal synergy: speaker information disambiguates visually similar faces and compensates for imperfect face detection, such as profile views, occlusion, or low resolution in archival footage. Conversely, face information helps when speaker embeddings are degraded by background noise, overlapping speech, or distance from the mic. The modest gain over face-only (0.4\,pp) is consistent with the unimodal asymmetry, as face embeddings already capture most of the discriminative signal for this corpus.

\textbf{Error analysis.} The remaining 1.4\,pp gap between the adaptive system and the oracle (95.5\% vs.\ 96.9\% on AVP) arises from misclassified queries. Two types of misclassification can occur: (a)~AoP/VoP queries misclassified as AVP, where the inactive modality contaminates fusion; and (b)~AVP queries misclassified as AoP/VoP, where a useful modality is discarded. Type~(a) is more harmful: it incurs a 3.9--4.9\,pp penalty per affected query, whereas type~(b) merely reduces the system to its strong face-only or speaker-only baseline.

This asymmetry explains the system's robustness. Even with ${\sim}$89\% detection accuracy, adaptive fusion yields substantial gains because correctly routing single-modality queries avoids large noise injection penalties, while the cost of missed multimodal opportunities remains comparatively small.

% ===========================================================================
\section{Conclusions}
\label{sec:conclusion}

We presented a query-adaptive framework that answers the question \textit{``to be multimodal or not to be''} for audio-visual person retrieval in uncurated broadcast archives. By detecting active modalities through cross-modal score consistency analysis, namely the agreement between one modality's retrieval set and the other's scores, the system achieves 94.2\% P@1 on BBC Rewind, outperforming all of the unimodal baselines and fixed fusion. The core finding is twofold: blind multimodal fusion can be counterproductive when one modality is absent, and cross-modal consistency provides a simple yet effective diagnostic to prevent it, a principle applicable to multimodal retrieval settings where modality presence cannot be guaranteed.

% ===========================================================================
\section{Acknowledgement}
\label{sec:ack}
This work was supported by the UK Engineering and Physical Sciences Research Council (EPSRC) under Grants EP/V002856/1, EP/V006223/1 and EP/V002740/2 (Multimodal Video Search by Examples), and by Cambridge University Press \& Assessment (CUP\&A), a department of the Chancellor, Masters, and Scholars of the University of Cambridge.
This work was conducted while the authors were affiliated with the institutions indicated by numerical superscripts in the author list; current affiliations are denoted by non-numerical symbols.
The authors also acknowledge the BBC for providing the data and the use of generative AI tools for proofreading purposes.

% ===========================================================================
\begin{comment}
\section{Acknowledgements}

This work was funded by the EPSRC under the MVSE project (EP/V002740/2). We thank the BBC for providing access to the Rewind corpus.
\end{comment}

% ===========================================================================
\bibliographystyle{IEEEtran}
\bibliography{mybib}

\end{document}